\documentclass[10pt,twocolumn,letterpaper]{article}

\usepackage[pagenumbers]{iccv} 

\def\method{LLaVA-KD}
\definecolor{gray_tab}{RGB}{235, 235, 235}
\definecolor{blue_tab}{RGB}{207, 220, 251}
\definecolor{lblu_tab}{RGB}{225, 235, 246}
\definecolor{oran_tab}{RGB}{252, 242, 237}
\definecolor{iccvblue}{rgb}{0.21,0.49,0.74}
\usepackage[pagebackref,breaklinks,colorlinks,allcolors=iccvblue]{hyperref}

\def\paperID{1925} 
\def\confName{ICCV}
\def\confYear{2025}

\title{\method: A Framework of Distilling Multimodal Large Language Models}

\author{%
    Yuxuan Cai$^{1}$\footnotemark[1]
  ~~ Jiangning Zhang$^{2,3}$\footnotemark[1]
  ~~ Haoyang He$^2$
  ~~ Xinwei He$^4$
  ~~ Ao Tong$^1$\\
  Zhenye Gan$^3$
  ~~ Chengjie Wang$^3$
  ~~ Zhucun Xue$^2$
  ~~ Yong Liu$^2$
  ~~ Xiang Bai$^1$$^{\dagger}$\\
    $^1$Huazhong University of Science and Technology ~~ $^2$Zhejiang University ~~ \\
  $^3$Youtu Lab, Tencent ~~ $^4$Huazhong Agricultural University\\
}

\begin{document}
\maketitle
\begin{abstract}
The success of Large Language Models (LLMs) has inspired the development of Multimodal Large Language Models (MLLMs) for unified understanding of vision and language. However, the increasing model size and computational complexity of large-scale MLLMs ($l$-MLLMs) limit their use in resource-constrained scenarios. Although small-scale MLLMs ($s$-MLLMs) are designed to reduce computational costs, they typically suffer from performance degradation.
To mitigate this limitation, we propose a novel \method~framework to transfer knowledge from $l$-MLLMs to $s$-MLLMs. 
Specifically, we introduce Multimodal Distillation (MDist) to transfer teacher model's robust representations across both visual and linguistic modalities, and Relation Distillation (RDist) to transfer teacher model's ability to capture visual token relationships.
Additionally, we propose a three-stage training scheme to fully exploit the potential of the proposed distillation strategy: 
\textit{1)} Distilled Pre-Training to strengthen the alignment between visual-linguistic representations in $s$-MLLMs, 
\textit{2)} Supervised Fine-Tuning to equip the $s$-MLLMs with multimodal understanding capacity, and 
\textit{3)} Distilled Fine-Tuning to refine $s$-MLLM's knowledge.
Our approach significantly improves $s$-MLLMs performance without altering the model architecture. Extensive experiments and ablation studies validate the effectiveness of each proposed component. Code will be available at \href{https://github.com/Fantasyele/LLaVA-KD}{https://github.com/Fantasyele/LLaVA-KD}.
\end{abstract}
    
\vspace{-2em}
\section{Introduction}
Inspired by the significant achievements of Large Language Models (LLM) in the field of Natural Language Processing, an emerging and rapidly developing research area is focusing on the development of Multimodal Large Language Models (MLLM). 
These models~\cite{llava15, bai2023qwen, yao2024minicpm} integrate visual encoder, feature projector, and LLM to achieve unified understanding of visual and linguistic information.
However, the success of MLLMs benefits from the scaling law, which significantly increases the model size. The large-scale model and high-cost inference limit the application of MLLMs in resource-constrained scenarios.

\begin{figure}
    \centering
    \includegraphics[width=0.85\linewidth]{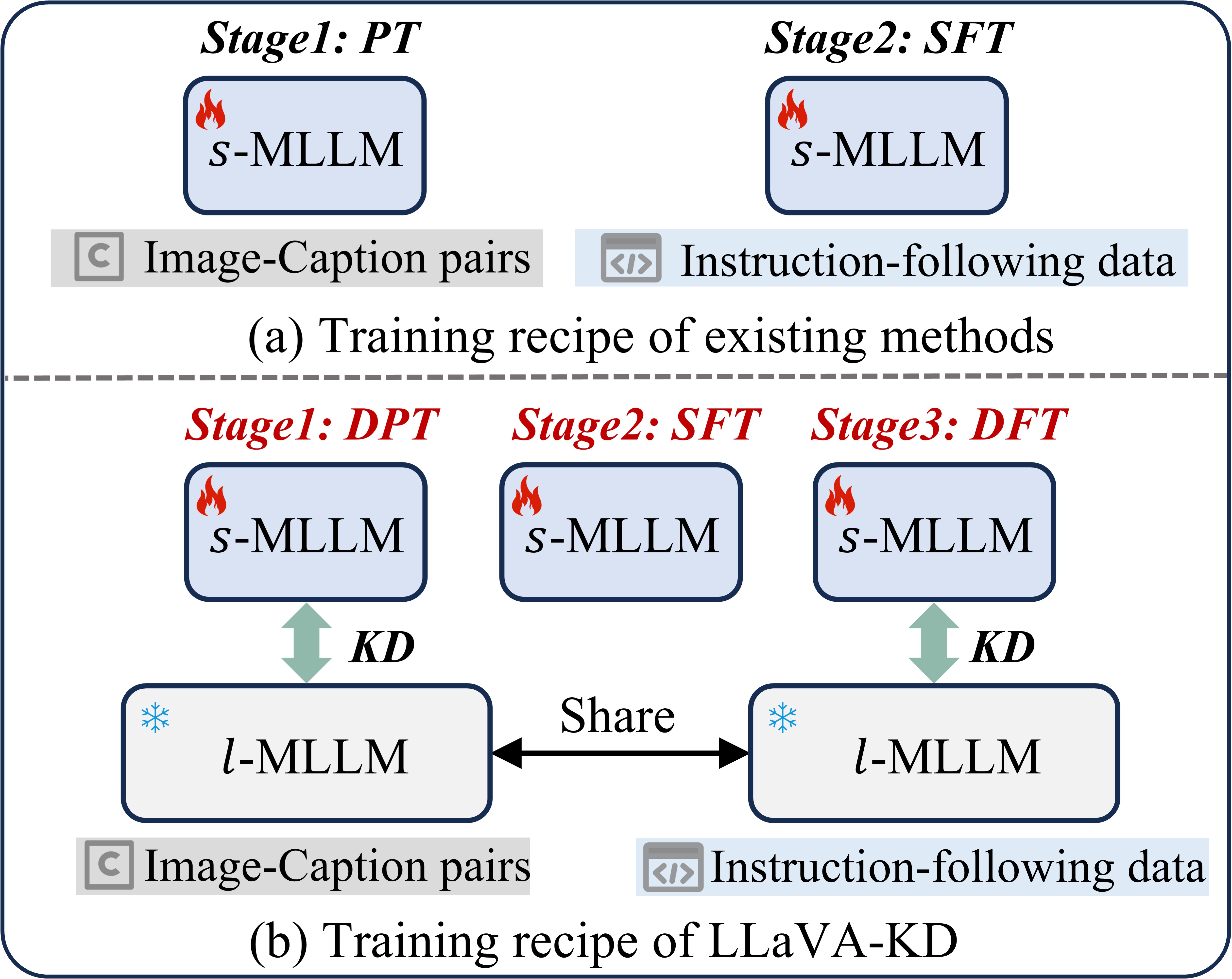}
    \vspace{-0.5em}
    \caption{To train a small-scale MLLM, (a) the existing methods follow a two-stage training scheme, including Pre-Training (PT) and Supervised Fine-Tuning (SFT). (b) Our LLaVA-KD proposes a three-stage scheme to exploit the potential of $s$-MLLM, including Distilled Pre-Training (DPT) to align visual-textual representation, SFT to equip the model with multimodal understanding, and Distilled Fine-Tuning (DFT) to refine $s$-MLLM's capacities.}
    \label{Fig:intro}
    \vspace{-1.4em}
\end{figure}

Some works~\cite{LLAVAphi, chu2023mobilevlm} address this challenge by directly adopting lightweight LLM backbones to build small-scale MLLMs ($s$-MLLMs), while preserving the conventional two-stage paradigm of large MLLMs ($l$-MLLMs) for model training, as illustrated in Fig.~\ref{Fig:intro}(a). This paradigm comprises two stages: Pre-Training (PT) and Supervised Fine-Tuning (SFT).
The PT stage establishes cross-modal alignment by projecting visual features to the text embedding space, while the SFT stage enhances multimodal understanding capabilities. Nevertheless, directly applying this paradigm-originally designed for $l$-MLLMs to $s$-MLLMs often yields suboptimal performance due to inherent capacity limitations~\cite{scalelaw}, preventing $s$-MLLMs from effectively learning the complex knowledge that $l$-MLLMs can capture.
Recent efforts compensate for this issue by optimizing model structure and improving the quality of training data: 
MoE-LLaVA~\cite{llavamoe} introduces the Mixture-of-Experts~\cite{moe} (MoE) to enhance the model's ability for understanding complex multimodal information while maintaining the computational cost, and Bunny~\cite{bunny} improves the training data quality by removing redundant data. 
Unlike these methods, we propose investigating training paradigm optimization as a novel pathway to enhance $s$-MLLMs performance, without architectural modifications or data engineering.

Knowledge distillation (KD) transfers knowledge from powerful teacher models to small student networks, significantly improves the performance of students with less computation. Although KD has achieved remarkable success in traditional vision tasks, its application to MLLMs remains underexplored.
Given that MLLMs primarily rely on language models to achieve robust cross-modal understanding, they present a challenge for distillation: \textit{how to effectively transfer multimodal understanding capabilities via language model distillation.} 
However, existing LLM distillation research focuses solely on knowledge transfer in the text modality, 
promoting the student model to mimic teacher-generated responses. Visual modality, however, play an equally crucial role in achieving comprehensive cross-modal understanding. To address this limitation, we propose two key innovations: (1) \textbf{Multimodal Distillation} (MDist), which extends distillation optimization to both visual and linguistic modalities, ensuring a holistic transfer of multimodal representations, and (2) \textbf{Relation Distillation} (RDist), which is predicated on the observation that structural relationships among visual tokens encode essential spatial and semantic dependencies for accurately understanding complex visual scenes, such as object positioning and inter-object interactions. We propose RDist to explicitly transfer the teacher model's capability to capture visual token relationships to the $s$-MLLM.

A straightforward approach is to introduce the above distillation strategy during the SFT stage to enhance $s$-MLLM multimodal understanding capacity. However, we observe that this approach exhibits limited benefits.
In this paper, motivated by two key observations, we propose an optimized three-stage training framework, as displayed in Fig.~\ref{Fig:intro}(b).
\textbf{First}, effective cross-modal alignment between visual and textual representations forms the foundation for multimodal understanding. To strengthen this alignment, we incorporate the Distillation during the PT stage (\textit{DPT}), enabling the $s$-MLLM to optimize the visual-linguistic joint embedding space through emulating the multimodal supervision signals from the $l$-MLLM.
\textbf{Second},
employing knowledge distillation exclusively during SFT is insufficient for comprehensive capability transfer. We therefore propose a \textit{SFT-DFT} (Supervised Fine-Tuning followed by Distilled Fine-Tuning) paradigm where the $s$-MLLM first establishes baseline understanding through standard SFT, then undergoes dedicated DFT to refine its knowledge.

Our method surpasses existing $s$-MLLM approaches across various multimodal benchmarks. We summarize our contributions as follows:

\begin{itemize}
\item We introduce LLaVA-KD, a novel MLLM-oriented distillation framework to transfers the knowledge from large-scale MLLM to the small-scale MLLM. Specifically, it contains a three-stage training scheme, including Distilled Pre-Training (DPT) for enhanced multimodal alignment, Supervised Fine-Tuning (SFT) for knowledeg acquisition, and Distilled Fine-Tuning (DFT) for comprehensive knowledge refinement.
\item We propose an innovative distillation strategy that integrates Multimodal Distillation (MDist) to optimize both visual and linguistic modalities with Relational Distillation (RDist), enhancing the $s$-MLLM's ability to capture relationships among visual tokens.
\item  Comprehensive evaluations across popular multimodal benchmarks demonstrate LLaVA-KD's superiority over recent $s$-MLLM advances.
\end{itemize}

\section{Related Works}
\subsection{Multimodal Large Language Model}
The emergence of large language models (LLMs) has promoted significant advances in MLLM for vision-language understanding. 
Early approaches like BLIP-2~\cite{BLIP2} employs a querying transformer architecture to establish cross-modal alignment for multimodal information understanding. In contrast, Flamingo~\cite{alayrac2022flamingo} introduces gated cross-attention layers to fuse visual features into LLMs.
Recent methods~\cite{llava,llava15, bai2023qwen} typically employ the lightweight projector such as Multi-Layer Perceptron (MLP) to align visual features with the text embedding space, followed by supervised instruction tuning on large-scale instruction dataset to enhance MLLMs's understanding capabilities.  
Most recently, a prominent research direction aims to enhance MLLMs' fine-grained visual perception through high-resolution input processing~\cite{minigemini,luo2024feast}, enabling diverse downstream applications including image segmentation~\cite{wu2025visionllm,zhang2025omg, zhang2024psalm} and grounding~\cite{zhang2024llava,ma2024groma}. 
Despite these advancements, the large model size and computational cost of current MLLMs greatly limit their application in resource-constrained scenarios, such as mobile devices.

\vspace{-1em}
\paragraph{Lightweight Multimodal Large Language Model.}
Existing lightweight MLLMs primarily reduce model parameters by employing lightweight LLM backbones. For example, LLaVA-Phi~\cite{LLAVAphi} adapts the LLaVA1.5~\cite{llava15} framework by substituting its foundation model with the parameter-efficient Phi-2 backbone. 
Recent studies demonstrate that optimizing model structure and training data can compensate for performance degradation caused by reduced model capacity: 
MoE-LLaVA~\cite{llavamoe} incorporates mixture-of-experts (MoE) into LLMs, achieving competitive multimodal understanding with merely 3B activated parameters.
Bunny~\cite{bunny} curates training data by clustering image embeddings via K-Means, followed by graph-based similarity pruning to reduce data size while maintaining diversity.

Unlike these methods, our approach focuses on improving the training paradigm for MLLMs.  In this paper, we propose a three-stage training framework based on knowledge distillation. By transferring the knowledge of large MLLMs to lightweight MLLMs, the light MLLMs' capabilities will be significantly enhanced.

\vspace{-0.5em}
\subsection{Knowledge Distillation}
\vspace{-0.3em}

Knowledge Distillation (KD)~\cite{hinton2015distilling} aims to transfer the knowledge from a large, complex teacher model to a lightweight, simple student model. This technique can significantly improve the performance of small models with fewer parameters, less computation, and faster speed. Knowledge distillation has been successful applied in visual tasks and has achieved success in many fields, typically in the domain of image classification. For example, 
Early work~\cite{hinton2015distilling} establishes soft target distillation using teacher logits as supervisory signals. Subsequent innovations like DKMF~\cite{kd_ic1} and FNKD~\cite{kd_ic2} reveal that intermediate feature alignment between teacher and student architectures yields superior classification accuracy.
DGKD~\cite{kd_ic3} further improves the student model's predictions by integrating multiple teacher models for guidance.

\vspace{-1.3em}
\paragraph{KD for LLM.} 

With the successful release of ChatGPT and its significant application value, LLM has gradually attracted attention and achieved numerous research progress in recent years~\cite{brown2020language,achiam2023gpt}. However, to achieve better results, the model size has also become increasingly larger which follows scaling law~\cite{scalelaw}, limiting its application in resource-constrained scenarios. This has promoted research into KD techniques for compressing LLMs while preserving their emergent capabilities.

MiniLLM~\cite{gu2024minillm} and DistiLLM~\cite{ko2024distillm} optimize the distillation process, proposing reverse Kullback-Leibler divergence (KLD) and skewed KLD objectives respectively, mitigating student overfitting to teacher distribution tails.
Wu~\textit{et al.}~\cite{wu2024rethinking} introduce an adaptive strategy that dynamically balances the weighting between standard KLD and reverse KLD objectives.
Parallel efforts~\cite{hsieh2023distilling,tian2024tinyllm,ranaldi2024aligning} leverage the Chain-of-Thought (CoT) capability of large LLMs to establish causal dependencies, thereby enriching training data. 

\vspace{-1.2em}
\paragraph{KD for MLLM.} 
Most recently, LLAVA-MoD~\cite{llavamod} advances knowledge distillation for training $s$-MLLMs. The method first optimizes the structure of the $s$-MLLM by integrating MoE~\cite{moe,llavamoe} into the LLM, thereby enhancing the model's expressive ability. During training, it firstly employs standard KLD to align the output distribution between the $s$-MLLM and $l$-MLLM. Additionally, it introduces a preference distillation process to improve the $s$-MLLM's discriminative capability, thereby reducing hallucinations.
LLaVADI~\cite{xu2024llavadi} is another $s$-MLLM work based on distillation, which reveals that most training strategies designed for LLMs do not bring additional benefits to the MLLMs,  suggesting the need for specialized optimization approaches in $s$-MLLM development. 

Unlike existing LLM/MLLM distillation methods that rely on complex constraint design, or elaborate architectural modifications, we focus on optimizing training schemes and developing multimodal distillation strategies. This enables efficient performance improvement for $s$-MLLMs while maintaining architectural simplicity.
\section{\method} \label{sec:method}
\begin{figure*}
    \centering
    \includegraphics[width=0.95\linewidth]{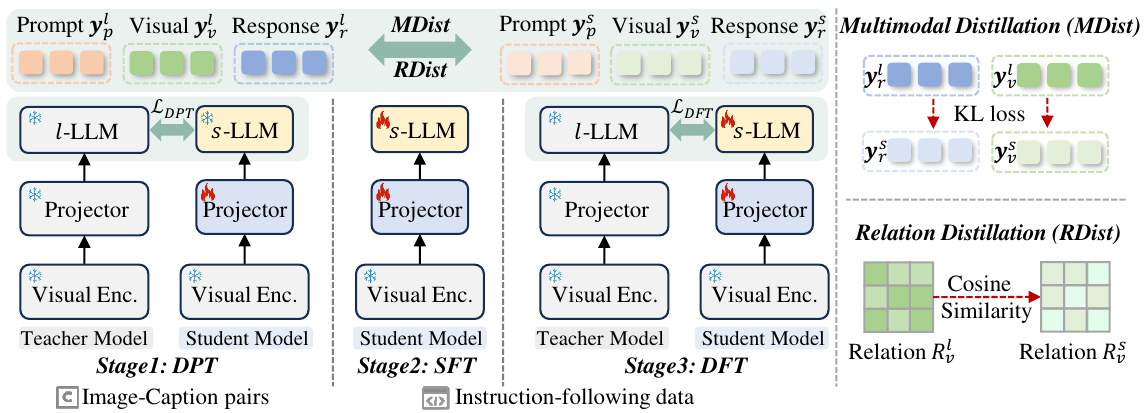} 
    \vspace{-0.6em}
    \caption{
    \textbf{Overview of our \method~} that contains three stages for effect training: 
    \textbf{\textit{1)}} Distilled Pre-Training (DPT) to align visual and text information as $l$-MLLM. 
    \textbf{\textit{2)}} Supervised Fine-Tuning (SFT) to enable $s$-MLLM with multimodal understanding capacity. 
    \textbf{\textit{3)}} Distilled Fine-Tuning (DFT) to refine $s$-MLLM's capacity by transferring $l$-MLLM's knowledge.
    During the training phase, we employ Multimodal Distillation (MDist) and Relation Distillation (RDist) in both DPT and DFT stages.}
    \label{fig:llavakd}
    \vspace{-1em}
\end{figure*}

The deployment of lightweight MLLMs is crucial for resource-constrained environments. However, small-scale MLLMs trained using conventional paradigm often yield suboptimal results. For instance,  while a 4B TinyLLaVA model achieves 65.0\% performance,reducing the model size to 0.5B results in a significant performance drop to 54.7\%. To mitigate this issue, we propose an innovative three-stage training scheme, incorporating a novel distillation strategy, termed \method, as illustrated in Fig.~\ref{fig:llavakd}.

\subsection{Composition of Distilled MLLM Architecture} \label{sec:composition} 
Fig.~\ref{fig:llavakd}(Left) illustrates the distillation process for MLLM, which includes a large-scale $l$-MLLM as the teacher model and a small-scale $s$-MLLM as the student model. Both them follow the architecture design of LLaVA-1.5~\cite{llava15}, and each includes three main components: \\
\noindent\textbf{Frozen Visual Encoder} is used to obtain robust visual features.
Specifically, given an input image $X_v \in \mathbb{R}^{H \times W \times 3}$, it is first sequenced to 2D patches $P_v \in \mathbb{R}^{N_p \times S_p^2 \times 3}$ with $S_p$ and $N_p$ representing patch size and its number, respectively. The subsequent transformer layers will project these patches to visual features $Z_v \in \mathbb{R}^{N_p \times C}$,  where $C$ denotes the dimension. Both teacher and student models use the same visual encoder, and we employ the pre-trained SigLIP~\cite{siglip} following previous success~\cite{bunny,cambrain}. 

\noindent\textbf{Visual Projector} contains two MLP layers with a GELU activation function to project visual features $Z_v$ into the text embedding space $H_v \in \mathbb{R}^{N_p \times D}$, where $D$ denotes the embedding dimensions.

\noindent\textbf{Large Language Model (LLM)} The LLM integrates multimodal information through joint processing of visual embeddings $H_v$ and text embeddings $H_t$. These representations are concatenated as $H = [H_v, H_t]$ and processed to generate sequential outputs: $\mathbf{y} =  [\mathbf{y}_p, \mathbf{y}_v, \mathbf{y}_r] = \{y_t\}_{t=1}^T$, 
where $\mathbf{y}_p$, $\mathbf{y}_v$, and $\mathbf{y}_r$ respectively denote prompt, visual, and response token sequences, and $T$ denotes the length of all prediction tokens.

\subsection{Training Scheme of Teacher Model $l$-MLLM} \label{sec:lmllm}
This section presents the common training scheme for powerful $l$-MLLMs, establishing performance upper bounds for $s$-MLLMs. This scheme consists of two stages: \\
\noindent\textbf{Pre-Training.} The \textit{Visual Encoder} and \textit{$l$-LLM} are kept frozen, and only the \textit{Projector} is optimized to align visual features with textual features. The training process employs image-caption pairs with the objective:
\begin{equation}
    \label{eq:autoreg}
    \mathcal{L}_{reg}=-\sum_{m=1}^{M} \log \phi_l \left(y_{m} \mid \mathbf{y}_{<m}\right),
\end{equation}
where $M$ denotes the length of predicted response tokens, while $\phi_l \left(y_{m} \mid \mathbf{y}_{<m}\right)$ denotes the distribution for the response token $y_m$ based on previous predictions $\mathbf{y}_{<m}$. \\
\noindent\textbf{Supervised Fine-Tuning.} 
This stage keeps the \textit{Visual Encoder} frozen, aiming at jointly optimizing \textit{Projector} and \textit{$l$-LLM} to enhance understanding and instruction-following capacities of the teacher model. 
Training employs high-quality instruction datasets with the objective function $\mathcal{L}_{SFT}$  following the autoregressive training paradigm defined in Equation~\ref{eq:autoreg}.

\vspace{-0.3em}
\subsection{Framework of LLaVA-KD} \label{sec:llavakd}
\vspace{-0.5em}
We first adopt the previous training strategy  (Sec.~\ref{sec:lmllm}) to develop the $l$-MLLM. 
For training $s$-MLLM, we propose a novel distillation strategy tailored for 
multimodal information learning (Sec.~\ref{sec:strategy}), and we further design a three-stage training scheme (Sec.~\ref{sec:threestage}).

\subsubsection{MLLM-Oriented KD Strategy}\label{sec:strategy}
\paragraph{Multimodal Distillation (MDist).} 
Considering that MLLM essentially leverages LLM for multimodal information understanding and reasoning, we follow the naive distillation method of LLM that uses standard Kullback-Leibler Divergence (KLD) to distill the response predictions. The training objective can be defined as:
\begin{equation}
\label{eq_kld_response}
\begin{aligned}
        \mathcal{L}_{res}  &= \sum_{m=1}^{M} \text{KLD}(\phi_{l}(y_m \mid \mathbf{y}_{<m}), \phi_{s}(y_m \mid \mathbf{y}_{<m})), \\
        &= \sum_{m=1}^{M} \sum_{j=1}^{V} \phi_{l}\left(Y_{j} \mid \mathbf{y}_{<m}\right) \log \left(\frac{\phi_{l}\left(Y_{j} \mid \mathbf{y}_{<m}\right)}{\phi_{s}\left(Y_{j} \mid \mathbf{y}_{<m}\right)}\right),
\end{aligned}
\end{equation}
where $M$ represents the length of response tokens and $V$ denotes the vocabulary space.
$\phi_{l}$ and $\phi_{s}$ denote the model parameters of $l$-MLLM and $s$-MLLM, respectively, $\phi_{l}\left(Y_{j} \mid \mathbf{y}_{<m}\right)$ and $\phi_{s}\left(Y_{j} \mid \mathbf{y}_{<m}\right)$ denote the probability of vocabulary $Y_j$ in the response token $y_m$, as predicted by $l$-MLLM and $s$-MLLM.

Meanwhile, the visual representation also plays a critical role for multimodal understanding of LLM. To leverage this property, we further optimize the output distributions of visual tokens from the teacher and student: 
\begin{equation}
\label{eq_kld_visual}
    \mathcal{L}_{vis} =\sum_{k=1}^{K} \sum_{j=1}^{V} \phi_{l}\left(Y_{j} \mid \mathbf{y}_{<k}\right) \log \left(\frac{\phi_{l}\left(Y_{j} \mid \mathbf{y}_{<k}\right)}{\phi_{s}\left(Y_{j} \mid \mathbf{y}_{<k}\right)}\right),
\end{equation}
where $K$ denotes the length of visual tokens, $V$ denotes the vocabulary space, $\phi_{l}\left(Y_{j} \mid \mathbf{y}_{<k}\right)$ and $\phi_{s}\left(Y_{j} \mid \mathbf{y}_{<k}\right)$ represent the probability of vocabulary $Y_j$ in the visual token $y_k$, as predicted by $l$-MLLM and $s$-MLLM respectively.

\vspace{-1em}
\paragraph{Relation Distillation (RDist).}
To empower the student model with fine-grained relational reasoning capabilities for complex visual scenes understanding, we construct a self-correlation matrix from the LLM-generated visual tokens. 
Through knowledge distillation by minimizing the matrix divergence, the student model inherits the teacher model's ability to comprehend the intricate relationships among visual tokens. 
To achieve this, we first compute the self-correlation matrices as follows:
\begin{equation}
\begin{cases}
\begin{aligned}
\label{eq_rela_matrix}
    R^s_v = \mathbf{y}^s_v \otimes \mathbf{y}^s_v \in \mathbb{R}^{N_p \times N_p}, \\
    R^t_v = \mathbf{y}^t_v \otimes \mathbf{y}^t_v \in \mathbb{R}^{N_p \times N_p},
\end{aligned}
\end{cases}
\end{equation}
where $\otimes$ represents matrix multiplication, 
$\mathbf{y}^s_v$ and $\mathbf{y}^t_v$ denote the visual tokens of the student and teacher, and $N_p$ denotes the number of visual tokens.
Subsequently, we maximum the cosine similarity between $R^s_v$ and $R^t_v$ that is defined as:
\begin{equation}
\label{eq_loss_rela}
    \mathcal{L}_{rel} = 1 - \text{Cos}(R^s_v, R^t_v) = 1 - \frac{R^s_v \cdot R^t_v}{\|R^s_v\| \|R^t_v\|},
\end{equation}
where $\text{Cos}(\cdot)$ denotes the cosine similarity.

\begin{table*}[]
\centering
\renewcommand{\arraystretch}{0.9}
\resizebox{1.\linewidth}{!}{
\begin{tabular}{cccccccc|ccccccc}
\toprule[0.17em]
\multirow{2}{*}{\textbf{Method}} & \multirow{2}{*}{LLM} & \multirow{2}{*}{\#Samples} & \multicolumn{5}{c}{Image Question Answering} & \multicolumn{5}{c}{Benchmarks}     & \multirow{2}{*}{$Avg_7$} & \multirow{2}{*}{$Avg_{10}$} \\\cline{4-13}
                                 &                      &                            & VQAv2  & GQA   & VisWiz  & SciQA  & TextVQA  & MME  & MMB  & $\text{MMB}^{\text{CN}}$ & POPE & MMMU &                      &                      \\\hline
LLaVA-1.5                        & Vicuna-7B            &        1.2 M                    & 78.5   & 62.0  & 50.0    & 66.8   & 58.2     & 75.5 & 64.3 & 58.3   & 85.9 & 34.4 & 62.2                 & 63.4                 \\
InstructBLIP                     & Vicuna-7B            &    130 M                        & -      & 49.2  & -       & 60.5   & 50.1     & -    & 36.0 & -      & -    & -    & -                    & -                    \\
Qwen-VL                          & Qwen-7B              &   1500 M                         & 78.8   & 59.3  & 35.2    & 67.1   & 63.8     & -    & 38.2 & 7.4    & -    & -    & -                    & -                    \\
Qwen-VL-Chat                     & Qwen-7B              &    1500 M                      & 78.2   & 57.5  & 38.9    & 68.2   & 61.5     & 74.4 & 60.6 & 56.7   & -    & 35.9 & 59.7                 & -                    \\
mPLUG-Owl2                       & LLaMA2-7B            &     400 M                       & 79.4   & 56.1  & 54.5    & 68.7   & 54.3     & 72.5 & 66.5 & -      & 85.8 & 32.7 & 62.1                 & -                    \\\hline
$\text{TinyLLaVA}^\dag$                        & Qwen1.5-4B           & 1.2 M                      & 79.9   & 63.4  & 46.3    & 72.9   & 59.0     & 69.3 & 67.9 & 67.1   & 85.2 & 38.9 & 63.7                 & 65.0                 \\
$\text{TinyLLaVA}^\dag$                        & Qwen2.5-3B           & 1.2 M                      & 80.4   & 63.2  & 38.7    & 76.0   & 61.5     & 73.9 & 71.8 & 69.5   & 86.4 & 40.3 & 64.9                 & 66.2                 \\
TinyLLaVA                        & Phi2-2.7B            & 1.2 M                      & 79.9   & 62.0  & -       & 69.1   & 59.1     & 73.2 & 66.9 & -      & 86.4 & 38.4 & -                    & -                    \\
Bunny                            & Phi2-2.7B            & 2.6 M                      & 79.8   & 62.5  & 43.8    & 70.9   & 56.7     & 74.4 & 68.6 & 37.2   & -    & 38.2 & 59.2                 & -                    \\
Imp-3B                           & Phi2-2.7B            & 1.5 M                      & -      & 63.5  & 54.1    & 72.8   & 59.8     & -    & 72.9 & 46.7   & -    & -    & -                    & -                    \\
MobileVLM                        & MLLaMA-2.7B          &   1.2 M                         & -      & 59.0  & -       & 61.0   & 47.5     & 64.4 & 59.6 & -      & 84.9 & -    & -                    & -                    \\
MoE-LLaVA                        & Phi2-2.7B            & 2.2 M                      & 79.9   & 62.6  & -       & 70.3   & 57.0     & -    & 68.0 & -      & 85.7 & -    & -                    & -                    \\
MiniCPM-V                        & MiniCPM-2.4B         &      570 M                      & -      & 51.5  & 50.5    & 74.4   & 56.6     & 68.9 & 64.0 & 62.7   & 79.5 & -    & 61.2                 & -                    \\
\rowcolor{oran_tab}
LLaVADI                          & MLLaMA-2.7B          &   1.2 M                         & -      & 61.4  & -       & 64.1   & 50.7     & 68.8 & 62.5 & -      & 86.7 & -    & -                    & -                    \\\hline
Imp-2B                           & Qwen1.5-1.8B         & 1.5 M                      & \underline{79.2}   & 61.9  & 39.6    & 66.1   & 54.5     & 65.2 & 63.8 & 61.3   & 86.7 & -    & 58.9                 & -                    \\
Bunny-2B                         & Qwen1.5-1.8B         & 2.6 M                      & 76.6   & 59.6  & 34.2    & 64.6   & 53.2     & 65.0 & 59.1 & 58.5   & 85.8 & -    & 56.3                 & -                    \\
Mini-Gemini-2B                   & Gemma-2B             & 2.7 M                      & -      & 60.7  & 41.5    & 63.1   & 56.2     & 67.0 & 59.8 & 51.3   & 85.6 & 31.7 & 57.1                 & -                    \\
MoE-LLaVA-2B                     & Qwen-1.5-1.8B        & 2.2 M                      & 76.2   & 61.5  & 32.6    & 63.1   & 48.0     & 64.6 & 59.7 & 57.3   & \textbf{87.0} & -    & 55.3                 & -                    \\
\rowcolor{gray_tab}
$\text{TinyLLaVA}^\dag$                        & Qwen2.5-1.5B         & 1.2 M                      & 78.8   & 62.0  & 43.2    & \textbf{72.0}   & 57.4     & \textbf{72.5} & 68.6 & 63.0   & 85.5 & 37.0 & 62.7                 & 64.0                 \\
\rowcolor{gray_tab}
$\text{TinyLLaVA}^\dag$                        & Qwen1.5-1.8B         & 1.2 M                      & 73.1   & 55.5  & 34.9    & 65.3   & 47.7     & 61.2 & 57.1 & 55.5   & 83.4 & 34.1 & 53.9                 & 56.8                 \\
\rowcolor{oran_tab}
LLaVA-MOD                        & Qwen1.5-1.8B         & 5 M                        & -      & 58.7  & 39.2    & 68.0   & \underline{58.5}     & 66.7 & 66.3 & 61.9   & \textbf{87.0} & -    & 59.9                 & -                    \\
\rowcolor{lblu_tab}
\method                              & Qwen1.5-1.8B         & 1.2 M                      & 79.0   & \underline{62.3}  &  \underline{44.7}    & 64.7   & 53.4     & 69.1 & 64.0 & 63.7   & 86.3 & 33.6 & 60.3                 & 62.1                 \\ 
\rowcolor{lblu_tab}
\method                              & Qwen2.5-1.5B         & 1.2 M                      & \textbf{80.3}   & \textbf{62.5}  & \textbf{46.0}    & \underline{71.6}   & \textbf{59.7}     & \textbf{70.0} & \textbf{71.0} & \textbf{66.6}   & 86.7 & \textbf{35.8} & \textbf{63.9}                 & \textbf{65.0}                 \\\hline
SPHINX-Tiny                      & TinyLlama-1.1B       & 15 M                       & 74.7   & 58.0  & 49.2    & 21.5   & 57.8     & 63.1 & 52.3 & 56.6   & 82.2 & -    & 51.2                 & -                    \\
\rowcolor{gray_tab}
$\text{TinyLLaVA}^\dag$                        & Qwen1.5-0.5B         & 1.2 M                      & 73.9   & 57.4  & 24.9    & 60.9   & 47.4     & 59.8 & 55.0 & 52.4   & 83.7 & 31.6 & 51.1                 & 54.7                 \\
\rowcolor{gray_tab}
$\text{TinyLLaVA}^\dag$                        & Qwen2.5-0.5B         & 1.2 M                      & 74.8   & 58.3  & 28.9    & 59.1   & 49.2     & 61.5 & 58.9 & 54.2   & 86.1 & 33.6 & 52.9                 & 56.5                 \\
\rowcolor{oran_tab}
LLaVADI                          & MLLaMA-1.4B          &       1.2 M                     & -      & 55.4  & -       & 56.0   & 45.3     & 58.9 & 55.0 & -      & 84.7 & -    & -                    & -                    \\
\rowcolor{oran_tab}
LLaVA-MOD                        & Qwen1.5-0.5B         & 5 M                        & -      & 56.2  & 31.6    & 62.8   & 53.9     & 65.3 & 58.8 & 50.4   & -    & -    & 54.1                 & -                    \\
\rowcolor{lblu_tab}
\method                              & Qwen1.5-0.5B         & 1.2 M                      & 77.0   & 59.6  & 35.9    & 60.6   & 49.9     & 64.5 & 60.1 & 55.5   & 85.9 & 30.2 & 55.2                 & 57.9                 \\
\rowcolor{lblu_tab}
\method                             & Qwen2.5-0.5B         & 1.2 M                      & \textbf{77.7}   & \textbf{59.8}  & 41.5    & 60.6   & 52.0     & 64.7 & \textbf{61.3} & \textbf{57.0}   & \textbf{86.4} & 28.3 & \textbf{56.7}                 & \textbf{58.9}  \\
\toprule[0.12em]
\end{tabular}
}
\vspace{-0.5em}
\caption{
\textbf{Benchmarked results with SoTA MLLMs}. Compared with counterparts, our \method~achieves highly competitive results than current small-scale MLLM models.
Optimal and sub-optimal results are in \textbf{bold} and \underline{underline}. 
\protect\sethlcolor{gray_tab}\hl{grey}, \protect\sethlcolor{oran_tab}\hl{orange} and \protect\sethlcolor{lblu_tab}\hl{blue} backgrounds represent ours baseline, the most direct MLLM distillation methods and our approach, respevtively.
$Avg_7$: The average of the seven benchmarks for direct comparison with existing MLLM distillation methods, excluding VQAv2, POPE, MMMU. $Avg_{10}$: The average across all benchmarks for comprehensive comparison. $^\dag$: reproduced results using the official code.}
\label{table:main_results}
\vspace{-1.5em}
\end{table*}

\subsubsection{Three-Stage Distillation Scheme} \label{sec:threestage}

Motivated by two critical observations, we propose a novel three-stage training paradigm,  as illustrated in Figure~\ref{fig:llavakd}. 
First, effective cross-modal alignment between visual and textual representations serves as the foundation for comprehensive multimodal understanding. To achieve this, we introduce Distilled Pre-Training (DPT) that enhances visual feature quality through targeted knowledge transfer, thereby promoting robust visual-textual alignment.
Second, we observe that directly applying distillation during SFT yields limited benefits. To address this limitation, we propose a paradigm where the student model first develops baseline comprehension through standard SFT, then undergoes refinement through Distilled Fine-Tuning (DFT). This phased approach enables progressive knowledge acquisition and refinement and ultimately enhances the model's overall multimodal understanding capacity.

\vspace{-1em}
\paragraph{Distilled Pre-Training (DPT).}
The main purpose of this stage is to project visual features to the text embedding space. Previous methods~\cite{llava15,LLAVAphi} primarily achieve this by optimizing the autoregressive prediction process of LLM (Eq.~\ref{eq:autoreg}). 
In contrast, \method~ further incorporates the distillation strategy to enhance cross-modal alignment.

Specifically,the visual encoder and language model components from the $s$-MLLM remain frozen during training, with optimization exclusive to the projection module. MDist is utilized to minimize the discrepancy between the student and the teacher model in terms of the output distribution of visual and response, while RDist is incorporated to enable the student to learn from the teacher model's ability to handle complex visual interactions. The combined MDist and RDist further optimize the quality of visual features, promoting the alignment between projected visual features and textual embeddings.

Overall, 
in addition to optimizing the autoregressive prediction results, we also utilize a multimodal distillation and relation distillation procedure. The objective is defined as:
\begin{equation}
    \mathcal{L}_{DPT} = \mathcal{L}_{reg} + \alpha \mathcal{L}_{res} + \beta \mathcal{L}_{vis} + \gamma \mathcal{L}_{rel},
\end{equation}
where $\mathcal{L}_{reg}$ denotes the auto-regressive prediction loss, $\alpha, \beta,$ are weights for visual and response distillation in MDist, and $\gamma$ is the weight for RDist.
\vspace{-1.2em}
\paragraph{Supervised Fine-Tuning (SFT).}
At this stage, we follow the common SFT procedure of the  $l$-MLLM training phase (Sec.~\ref{sec:lmllm}). By jointly training the Projector and $s$-LLM, the model is initialized with multimodal information understanding ability and instruction-following capability. The training objective can be defined as Eq.~\ref{eq:autoreg}, denoted as $\mathcal{L}^\prime_{SFT}$.
\vspace{-2.2em}
\paragraph{Distilled Fine-Tuning (DFT).} 
The main objective of this stage is to further refine the capacity of the $s$-MLLM. 
Specifically, we integrate the distillation strategies of MDist and RDist during this training phase to effectively acquire knowledge from the $l$-MLLM. MDist enables the $s$-MLLM to directly emulate the $l$-MLLM's sophisticated multimodal understanding capabilities, whereas RDist enhances the $s$-MLLM's visual representations by learning to capture visual token relationships, promoting a more context-aware understanding in the $s$-MLLM.
Overall, the training objective can be defined as:
\begin{equation}
    \mathcal{L}_{DFT} = \mathcal{L}_{reg} + \alpha^\prime \mathcal{L}_{res} + \beta^\prime \mathcal{L}_{vis} + \gamma^\prime \mathcal{L}_{rel}, 
\end{equation}
where $\mathcal{L}_{reg}$ denotes the auto-regressive prediction loss, $\alpha^\prime, \beta^\prime$ are weights for visual and response distillation in MDist, and $\gamma^\prime$ is the weight for RDist.
\vspace{-0.3em}
\section{Experimental Results}
\subsection{Setup}
\vspace{-0.35em}
\paragraph{Implementation Details.}
For both the large/small-scale MLLMs, we employ the pre-trained SigLIP-B/14@384px~\cite{siglip} as the Visual Encoder and a two-layer MLP with a GELU activation layer as the Projector, while adopting Qwen1.5~\cite{yang2024qwen2} and Qwen2.5 families as LLM models. 
During training, we utilize the LLaVA1.5-558k~\cite{llava15} for the DPT stage, and LLaVA-mix-665k~\cite{llava15} for both SFT and DFT stages.
The loss weights ($\{\alpha, \beta, \gamma\}$ and $\{\alpha^\prime$, $\beta^\prime$, and $\gamma^\prime\}$ ) during DPT and DFT stages are set to 1.0, 1.0, and 0.5, respectively.  
The detailed training hyperparameter configurations are illustrated in Appendix.

\vspace{-1.2em}
\paragraph{Details of Comparison Methods.}
We primarily compare with recent efforts focused on small-scale MLLMs, including Imp~\cite{shao2024imp}, Bunny~\cite{bunny}, Mini-Gemini~\cite{minigemini}, MoE-LLaVA~\cite{llavamoe}, SPHINX~\cite{gao2024sphinx}, TinyLLaVA~\cite{tinyllava} and LLaVA-MoD~\cite{llavamod}. 
Additionally, we also compare our LLaVA-KD with current state-of-the-art MLLMs, such as Instruct-BLIP~\cite{instructblip}, mPLUG-Owl2~\cite{ye2024mplug}, LLaVA1.5~\cite{llava15}, Qwen-VL~\cite{bai2023qwen}, MobileVLM~\cite{chu2023mobilevlm}, MiniCPM-V~\cite{yao2024minicpm}.
\vspace{-1.2em}
\paragraph{Benchmark Settings.}
General VQA requires models to generate answers based on the image and related question, necessitating the ability to understand how visual and textual information interrelate. For general VQA, we evaluate LLaVA-KD on four benchmarks including VQAv2~\cite{vqa2}, GQA~\cite{gqa}, VizWiz~\cite{vizwiz}, and ScienceQA (Image set)~\cite{sqa}. 
Scene Text-centric VQA (TextVQA~\cite{textvqa}) requires the model recognize and understand textual information in an image. 
Additionally, we use five popular benchmarks for comprehensive evaluation including MME~\cite{mme}, MMB~\cite{mmb}, $\text{MMB}^{\text{CN}}$~\cite{mmb},  POPE~\cite{pope}, and MMMU~\cite{mmmu}.

\vspace{-0.35em}
\subsection{Benchmarked Results with the SoTAs}
\vspace{-0.35em}
As shown in Table~\ref{table:main_results}, 
our LLaVA-KD demonstrates superior performance across both 1B and 2B model scales. 
Specifically, for the 1B configuration, when initialized with Qwen1.5-0.5B,
LLaVA-KD achieves a 4.0\% average improvement in $Avg_7$ over SPHINX-Tiny~\cite{gao2024sphinx}, using only 1M training samples compared to SPHINX-Tiny's 15M.  Meanwhile, we outperform the recent MLLM distillation art LLaVA-MoD~\cite{llavamod} by 1.1\% in $Avg_7$, while requiring less training data (1.2M vs. 5M). Notably, LLaVA-MoD introduces a MoE structure in the $s$-MLLM, resulting in large total parameters.
Further employing Qwen2.5-0.5B  improves LLaVA-KD's performance.
Moreover,  compared to the TinyLLaVA baseline (Qwen1.5-0.5B/Qwen2.5-0.5B) using conventional PT-SFT training scheme, LLaVA-KD achieves improvements of 3.2\% and 2.4\% in $Avg_{10}$ respectively across all ten benchmarks.

Our advantage extends to the 2B model scale, achieving the leading performance compared to existing small-scale MLLMs. When built upon Qwen1.5-1.8B, our model outperforms both LLaVA-MoD~\cite{llavamod} and Imp-2B~\cite{shao2024imp} by 0.4\% and 1.4\% in $Avg_7$ respectively, while maintaining greater data efficiency. Similar to the 1B observations, LLaVA-KD-2B demonstrates comprehensive improvements over TinyLLaVA baseline (Qwen1.5-1.8B/Qwen2.5-1.5B) with gains of 5.3\% and 1.0\% in $Avg_{10}$, respectively.

\begin{table*}[]
    \centering
    \renewcommand{\arraystretch}{0.8}
    \setlength\tabcolsep{6.0pt}
    \resizebox{0.8\linewidth}{!}{
        \begin{tabular}{lccccc|cccccc}
        \toprule[0.17em]
        \multicolumn{1}{c}{\multirow{2}{*}{Training Scheme}} & \multicolumn{5}{c}{Image Question Answering}                                & \multicolumn{5}{c}{Benchmarks}                                                & \multirow{2}{*}{$Avg_{10}$} \\\cline{2-11}
        \multicolumn{1}{c}{}                                 & VQAv2       & GQA           & VizWiz        & SciQA         & TextVQA       & MME   & MMB   & MMB$^{\text{CN}}$    & POPE          & MMMU    &                      \\\hline
        PT-SFT                                               & 73.9        & 57.4          & 24.9          & 60.9          & 47.4          & 59.8          & 55.0            & 52.4          & 83.7          & \textbf{31.6} & 54.7                 \\
        
        DPT-SFT                                              & 74.6        & 57.8          & 28.8          & \textbf{61.2} & 49.1          & 59.9          & 56.9          & 51.6          & 84.3          & 31.4          & 55.6                \\
        PT-DFT & 75.1 & 57.0	&	29.5 & 60.9	&49.2		&59.6 &	57.3	&55.0 &85.5	&29.6 & 55.8\\
        DPT-DFT      & 75.5        & 58.0            & 27.5          & 59.7          & 49.3          & 60.6          & 57.7          & 54.7          & 85.4          & 30.3          & 55.9        \\
        PT-SFT-DFT & 76.6 & 59.4 & 32.6 & 60.4 & 48.4 & 60.9 & 57.8 & 54.0 & 84.9 & 31.3 & 56.6 \\
        DPT-SFT-DFT                                          & \textbf{77.0} & \textbf{59.6} & \textbf{35.9} & 60.6          & \textbf{49.9} & \textbf{64.5} & \textbf{60.1} & \textbf{55.5} & \textbf{85.9} & 30.2          & \textbf{57.9}     \\
        \textcolor{gray}{DPT-DFT-DFT}  & \textcolor{gray}{77.5}	& \textcolor{gray}{60.3}	& \textcolor{gray}{37.6}	& \textcolor{gray}{61.1}	& \textcolor{gray}{49.3}	& \textcolor{gray}{63.2} & \textcolor{gray}{59.4} & 	\textcolor{gray}{54.9} & \textcolor{gray}{86.0}			& \textcolor{gray}{31.0} & \textcolor{gray}{58.0} \\
        \toprule[0.12em]
        \end{tabular}
            }
        \vspace{-1.0em}
        \caption{Ablation studies of different training stages.}
        \label{table:ablation_stages}
        \vspace{-1.3em}
\end{table*}

\begin{table}
\centering
\renewcommand{\arraystretch}{0.5}
\setlength\tabcolsep{9.0pt}
\resizebox{0.8\linewidth}{!}{
\begin{tabular}{cccccc}
 \toprule[0.12em]
\multicolumn{2}{c}{DPT} & \multicolumn{1}{l}{\multirow{2}{*}{SFT}} & \multicolumn{2}{c}{DFT} & \multirow{2}{*}{$Avg_{10}$} \\\cline{1-2}
\cline{4-5}
MDist & RDist & \multicolumn{1}{l}{}                 & MDist & RDist      &                          \\
\hline 
\XSolidBrush  & \Checkmark & \multirow{3}{*}{\Checkmark}   & \XSolidBrush & \XSolidBrush & 55.5 \\ 
\Checkmark                     & \XSolidBrush                     &                                           & \XSolidBrush                          & \XSolidBrush                          & 55.1                     \\
\Checkmark                     & \Checkmark                     &                                                             & \XSolidBrush                          & \XSolidBrush                          & \textbf{55.6}                     \\
\hline
\Checkmark                     & \Checkmark  & \multirow{3}{*}{\Checkmark} & \XSolidBrush & \Checkmark  & 57.0 \\
\Checkmark                     & \Checkmark                     &                                           & \Checkmark                          & \XSolidBrush                          & 57.7                     \\
\Checkmark                     & \Checkmark                     &                                                             & \Checkmark                          & \Checkmark                          & \textbf{57.9}     \\
 \toprule[0.1em]
\end{tabular}
}
\vspace{-1.0em}
\caption{Ablation Study on MDist and RDist.}
\label{tab:ablation_c}
\vspace{-1em}
\end{table}

\begin{table}[]
\centering        
\renewcommand{\arraystretch}{0.7}
\setlength\tabcolsep{3.0pt}
\resizebox{0.8\linewidth}{!}{
    \begin{tabular}{ccccc}
    \toprule[0.12em]
    Training Stage & Response tokens     & Prompt tokens     & Visual tokens & $Avg_{10}$ \\
    \hline
     \multirow{4}{*}{DPT} & \Checkmark             & \XSolidBrush                & \XSolidBrush          & 54.9                     \\
    & \Checkmark             & \Checkmark                & \XSolidBrush          & 55.0                       \\
    & \Checkmark             & \XSolidBrush                & \Checkmark          & 55.1                     \\
    & \Checkmark             & \Checkmark                & \Checkmark          & 54.6           \\\hline
     \multirow{4}{*}{DFT} & \Checkmark             & \XSolidBrush                & \XSolidBrush          & 57.2                     \\
    &  \Checkmark             & \Checkmark                & \XSolidBrush          & 56.9                       \\
    & \Checkmark             & \XSolidBrush                & \Checkmark          & 57.7                     \\
    & \Checkmark             & \Checkmark                & \Checkmark          & 57.1           \\
     \toprule[0.1em]
    \end{tabular}
    }
\vspace{-0.9em}
\caption{Ablation Study on Distillation Targets.}
\label{table:ablation_all}
\vspace{-1.7em}
\end{table}

\vspace{-0.35em}
\subsection{Ablation Study and Analysis} 
\vspace{-0.35em}
\paragraph{Three-Stage Training Recipe.}
In Table~\ref{table:ablation_stages}, 
we investigate the impact of different training stages. Initially, we follow the conventional Pre-Training (PT) followed by Supervised Fine-Tuning (SFT) recipe to train the small MLLM, which serves as our baseline (Row 1) with an average performance of 54.7\%. 
Incorporating DPT alongside SFT (Row 2) results in a 0.9\% improvement (55.6\% vs. 54.7\%), suggesting that DPT enhances the alignment between visual feature and text embedding and thereby facilitates the LLM's understanding of multimodal information.
When we further employ DFT (DPT-SFT-DFT, Row 6), the model's performance improves significantly by 2.3\%, achieving 57.9\% in $Avg_{10}$. The result indicates that DFT effectively enables the $s$-MLLM to learn knowledge from the $l$-MLLM, substantially enhancing its understanding capabilities.
In contrast, omitting the SFT stage (DPT-DFT, Row 4) results in a notable performance drop to 55.9\%, underscoring the necessity of the SFT stage for knowledge acquisition. 
Notably, the DPT-DFT achieves optimal results in the two-stage training paradigm. Introducing an additional DFT phase (Row 7) yields comparable performance to DPT-SFT-DFT with 58.0\%, but requires increased computational overhead (120 GPU hours). This indicates that DPT-SFT-DFT achieves the best performance-computation trade-off.

Additionally, we observe that the performance of PT-DFT (Row 3) and DPT-DFT is comparable. Meanwhile, when the SFT stage is incorporated, the DPT-based pipeline (DPT-SFT-DFT, Row 6) outperforms PT-based pipeline (PT-SFT-DFT, Row 5) by 1.3\%, emphasizing the advantage provided by DPT. The comparable results obtained in the absence of SFT highlight the critical role of DFT, which enhances the model's comprehension ability regardless of the pretraining strategy. Although DPT establishes superior multimodal alignment during initial pretraining, DFT's robust knowledge transfer compensates for the limitations of PT, ultimately leading to similar performance between PT-DFT and DPT-DFT.

\vspace{-1em}
\begin{table*}[]
    \centering
    \renewcommand{\arraystretch}{0.8}
    \resizebox{0.8\linewidth}{!}{
    \begin{tabular}{cccccc|cccccc}
        \toprule[0.17em]
        \multicolumn{1}{c}{\multirow{2}{*}{Distillation strategy}} & \multicolumn{5}{c}{Image Question Answering}                                & \multicolumn{5}{c}{Benchmarks}                                                & \multirow{2}{*}{$Avg_{10}$} \\\cline{2-11}
        \multicolumn{1}{c}{}                                 & VQAv2       & GQA           & VizWiz        & SciQA         & TextVQA       & MME   & MMB   & MMB$^{\text{CN}}$    & POPE          & MMMU    &                      \\\hline
         FKL  & 74.3 & 56.1 & 31.7 & 59.4 & 49.0 & 58.9 & 57.4 & 54.0 & 84.4 & 29.8 & 55.5 \\
         RKL~\cite{gu2024minillm} & 74.3 & 56.6 & 26.7 & 60.8 & 49.1 & 57.8 & 56.8 & 53.7 & 84.7 & \textbf{30.0} & 55.0 \\
         JSD~\cite{jsd_ACL} & 73.8 & 54.9 & \textbf{32.3} & 60.3 & 48.7 & 57.6 & \textbf{57.8} & 54.3 & 85.1 & 29.8 & 55.5  \\
         Ours & \textbf{75.1} & \textbf{57.0} & 29.5 & \textbf{60.9} & \textbf{49.2} & \textbf{59.6} & 57.3 & \textbf{55.0} & \textbf{85.5} & 29.6 & \textbf{55.8} \\
         \toprule[0.12em]
    \end{tabular}
    }
    \vspace{-0.9em}
    \caption{Comparison with Distillation Strategies in LLMs.} 
    \label{tab_llm}
    \vspace{-1em}
\end{table*}

\begin{table*}[]
\centering
\renewcommand{\arraystretch}{0.8}
\resizebox{0.8\linewidth}{!}{
\begin{tabular}{cccccc|cccccc}
\toprule[0.17em]
\multirow{2}{*}{Distillation Strategy}                                  & \multicolumn{5}{c}{Image Question Answering} & \multicolumn{5}{c}{Benchmarks}                           & \multirow{2}{*}{$Avg_{10}$} \\\cline{2-11}
                                & VQAv2  & GQA   & VisWiz  & SciQA  & TextVQA  & MME & MMB &  MMB$^{\text{CN}}$ & POPE & MMMU &                      \\\hline
MD                            & 76.3   & 58.5  & 31.6    & 58.4   & \textbf{51.7}     & 60.6       & 59.6        & \textbf{55.8}       & 86.2 & 30.2      & 56.9               \\
MD+PD  & 74.4   & 57.1  & 22.7    & 58.4   & 47.7     & 58.4     & 58.8        & 54.5       & \textbf{86.6} & \textbf{32.1}      & 55.1              \\
Ours                                           & \textbf{77.0}     & \textbf{59.6}  & \textbf{35.9}    & \textbf{60.6}   & 49.9     & \textbf{64.5}       & \textbf{60.1}        & 55.5       & 85.9 & 30.2      & \textbf{57.9}    \\
\toprule[0.12em]
\end{tabular}
}
\vspace{-0.8em}
\caption{Comparison with distillation pipeline in LLaVA-MOD. MD and PD denote the Mimic Distillation and Preference Distillation.}
\label{tab_llavamod}
\vspace{-1.3em}
\end{table*}

\begin{table}[]
\centering
\renewcommand{\arraystretch}{0.7}
\resizebox{0.9\linewidth}{!}{
\begin{tabular}{cclc}
\toprule[0.17em]
LLM of the Teacher & LLM of the Student & Training Recipe & $Avg_{10}$ \\
\midrule
MLLaMA 2.7B                        & /                                 & PT-SFT                            & 55.2                                          \\
/                                 & MLLaMA 1.7B                     & PT-SFT                            & 48.8                                          \\
 MLLaMA 2.7B                       &  MLLaMA 1.7B                   & DPT-SFT                           &  50.5                     \\
MLLaMA 2.7B                        &  MLLaMA 1.7B                   & DPT-SFT-DFT                       & 53.4           \\
\toprule[0.12em]
\end{tabular}
}
\vspace{-0.9em}
\caption{Verification on MobileVLM.}
\vspace{-1.3em}
\label{tab:mvlm_frame}
\end{table}

\begin{table}[]
\centering
\renewcommand{\arraystretch}{0.8}
\resizebox{0.9\linewidth}{!}{
\begin{tabular}{cclc}
\toprule[0.17em]
LLM of the Teacher & LLM of the Student & Training Recipe & $Avg_{10}$ \\\hline
Qwen1.5-4B                                              & /                                                       & PT-SFT                           & 65.0                     \\
Qwen1.5-7B                                              & /                                                       & PT-SFT                           & 65.7                     \\
/                                                       & Qwen2.5-0.5B                                            & PT-SFT                           & 54.7                     \\
Qwen1.5-4B                                              & \multirow{2}{*}{Qwen1.5-0.5B}                           & DPT-SFT-DFT                      & 57.9                     \\
Qwen1.5-7B                                              &                                                         & DPT-SFT-DFT                      & 57.4                     \\\hline
Qwen2.5-3B                                              & /                                                       & PT-SFT                           & 66.2                     \\
Qwen2.5-7B                                              & /                                                       & PT-SFT                           & 69.3                     \\
/                                                       & Qwen2.5-0.5B                                            & PT-SFT                           & 56.5                     \\
Qwen2.5-3B                                              & \multirow{2}{*}{Qwen2.5-0.5B}                           & DPT-SFT-DFT                      & 58.9                     \\
Qwen2.5-7B                                              &                                                         & DPT-SFT-DFT                      & 58.3          \\
\toprule[0.12em]
\end{tabular}
}
\vspace{-0.9em}
\caption{Ablation study on teacher models with different sizes.}
\label{tab:teacher_scale}
\vspace{-1.3em}
\end{table}

\vspace{-0.5em}
\paragraph{Distillation Strategy.} \textcolor{black}{Table~\ref{tab:ablation_c} explores the influence of different distillation strategies, including MDist and RDist, during both the DPT and DFT stages.
First, we report the results of DPT using different distillation strategies, followed by SFT (Rows 1-3). 
The results show that using RDist alone is more effective than using MDist alone, with this improvement potentially attributed to RDist's effectiveness in preserving relationships in visual representations, thereby enhancing feature quality and promoting effective visual-textual alignment for improved cross-modal understanding in the $s$-MLLM.
During the DFT stage (Rows 4-6), MDist proves to be more effective than RDist. The advantage likely arises from the direct alignment of the output multimodal distribution of large MLLMs, which enhances the understanding capabilities of small MLLMs.
Across both distillation stages, combining MDist and RDist achieves optimal performance. The results demonstrate that integrating both MDist and RDist helps to comprehensively transfer the knowledge from large MLLMs to small MLLMs.}

\vspace{-1.4em}

\paragraph{Distillation Targets.}
\textcolor{black}{As shown in Table~\ref{table:ablation_all}, 
we validate the effectiveness of different distillation targets in MDist during both the DPT stage and DFT stage. 
The results indicate that, unlike most existing methods that focus solely on distilling the response, incorporating visual distillation achieves the best results, whether in the DPT or DFT stage.
This improvement can be attributed to that, in the DPT stage, adding visual constraints helps improve the quality of the projected visual space in the $s$-MLLM, thereby promoting the alignment of visual and language information, facilitating unified understanding by the LLM. 
Furthermore, during the DFT stage, direct imitation of visual-textual cross-modal comprehension in $l$-MLLM, further enhances the $s$-MLLM's understanding and reasoning capabilities.}

\vspace{-1.3em}
\paragraph{Comparison of Distillation strategy of LLM.} \textcolor{black}{We further evaluate our proposed MDist and RDist distillation strategy against existing LLM distillation approaches under the PT-DFT training recipe, as illustrated in Table~\ref{tab_llm}. Our method achieves the best results across seven benchmarks, demonstrating an average improvement of 0.3\% compared to Forward Kullback-Leibler Divergence (FKL) and Jensen–Shannon Divergence~\cite{jsd_ACL} (JSD), and an average improvement of 0.8\% compared to Reverse Kullback-Leibler Divergence~\cite{gu2024minillm} (RKL).}

\textcolor{black}{Unlike existing LLM distillation methods that primarily focus on the optimization of textual information, our distillation strategy incorporates the role of visual information in multimodal understanding tasks. 
This approach enables small-scale MLLMs to better align with the multimodal representations of large-scale MLLMs, as well as improving the ability to capture the relationships across visual tokens. By refining the distillation process to emphasize visual token alignment, the $s$-MLLM generates more context-aware predictions, leading to superior overall performance.}

\vspace{-1.3em}
\paragraph{Comparison with the distillation pipeline in LLaVA-MOD.}

\textcolor{black}{LLaVA-MOD~\cite{llavamod} introduces a combined distillation pipeline of Mimic Distillation (MD) and Preference Distillation (PD). MD enables the $s$-MLLM to emulate the understanding of the $l$-MLLM. In contrast, PD enhances the $s$-MLLM's discriminative capacity by leveraging the preference signals from the $l$-MLLM.}
\textcolor{black}{To compare with LLaVA-MOD, we replace our DFT stage in the DPT-SFT-DFT paradigm with MD and PD.
As illustrated in Table~\ref{tab_llavamod}, when employing the MD, our method demonstrates an average improvement of 1.0\%. Moreover, integrating PD alongside MD reduces overall performance to 55.1\%, with our method showing an advantage of 2.8\%.}
\textcolor{black}{This aligns with LLaVA-MoD's observation that PD primarily reduces the hallucination of $s$-MLLM, but this also results in a decline in the model's comprehension capability.}

\vspace{-1.3em}
\paragraph{Verification on another MLLM framework.}
\textcolor{black}{We additionlly conduct experiments on the MobileVLM~\cite{chu2023mobilevlm} framework to evaluate the generalization of our training algorithm, as illustrated in Table~\ref{tab:mvlm_frame}. The $l$-MLLM and $s$-MLLM employ MobileLLaMA-2.7B and -1.7B respectively, achieving 55.2\% and 48.8\% average performance across ten benchmarks. 
Application of DPT-SFT to $s$-MLLM yields 50.5\% performance. The complete DPT-SFT-DFT paradigm achieves 53.4\%, achieving 4.6\% improvement over conventional PT-SFT. Refer to Appendix for detailed results.}

\vspace{-1.3em}
\paragraph{Impact of teacher models with different sizes.} 

In Table~\ref{tab:teacher_scale}, we investigate the impact of teacher model size on LLaVA-KD by employing different variants of LLMs (Qwen1.5-4B/7B and Qwen2.5-3B/7B). The results demonstrate that our LLaVA-KD with either teacher variant significantly enhances the capabilities of $s$-MLLM compared to the conventional PT-SFT paradigm. However, counterintuitively, 
larger teachers (Qwen1.5-7B and Qwen2.5-7B) exhibit lower distillation effectiveness than the smaller ones (Qwen1.5-4B and Qwen2.5-3B).

This finding aligns with recent knowledge distillation studies~\cite{yang2024clip, busbridge2025distillation}, which suggest that beyond a critical capability threshold, increasing teacher capacity yields diminishing returns for student performance. 
Specifically, excessive capability gaps between teachers and students create an alignment bottleneck during distillation, limiting the student model's expressive capacity. 
This insight highlights the importance of optimizing teacher-student capacity matching in MLLMs knowledge distillation, suggesting an open issue for future research.

\vspace{-1.4em}
\paragraph{Futher Exploration.} \textcolor{black}{It should be noted that in our framework, 
to ensure that the $s$-MLLM can effectively learn from the $l$-MLLM, both $l$-MLLM and $s$-MLLM need to employ the same series of LLMs to maintain consistency in the vocabulary space. 
Future research could address both architectural limitations and capacity mismatch simultaneously, potentially enabling integration of heterogeneous MLLMs to leverage complementary knowledge sources. This dual-axis improvement pathway may yield more powerful teacher models while further enhancing $s$-MLLM performance through optimized knowledge transfer.
}

\vspace{-1em}
\section{Conclusion} 
\vspace{-0.3em}
This paper introduced LLaVA-KD, a framework that transfers knowledge from $l$-MLLM to $s$-MLLM. This approach effectively reduces model size and computational complexity while enabling the $s$-MLLM to maintain the capabilities.
\textcolor{black}{LLaVA-KD introduces a distillation strategy, including MDist and RDist. MDist simultaneously transfer the visual-textual cross-modal information, whereas RDist transfers $l$-MLLM's capacity to capture correlations among visual representations.}
Additionally, we propose a simple yet effective three-stage training framework to fully exploit the potential of distillation: DPT to promote the alignment between visual-textual representations in the $s$-MLLM, SFT to equip model with multimodal understanding, and DFT to further refine $s$-MLLM's capacity. Comprehensive experiments reveal the effectiveness of our framework.

\bibliographystyle{ieeenat_fullname}
\bibliography{main}

\end{document}